\documentclass{article}

\usepackage[final]{neurips_2024}
\usepackage[utf8]{inputenc} %
\usepackage[T1]{fontenc}    %
\usepackage{url}            %
\usepackage{booktabs}       %
\usepackage{amsfonts}       %
\usepackage{nicefrac}       %
\usepackage{microtype}      %
\usepackage[table]{xcolor}         %
\usepackage{multirow}
\usepackage{amsmath}
\usepackage{amssymb}
\DeclareMathOperator*{\argmax}{arg\,max}

\usepackage{subcaption}
\usepackage{diagbox}
\usepackage{graphicx}
\usepackage{wrapfig}
\usepackage{overpic}
\usepackage{soul}
\usepackage{algorithm}
\usepackage{algpseudocode}
\usepackage{makecell}
\usepackage[graphicx]{realboxes}
\usepackage{tikz}
\usepackage{hyperref}  
\usepackage[capitalize]{cleveref}   %
\Crefname{figure}{Fig.}{Figs.}
\Crefname{equation}{Eq.}{Eqs.}
\Crefname{section}{Sec.}{Secs.}
\newcommand{\eg}{\textit{e.g.}}
\newcommand{\ie}{\textit{i.e.}}

\title{OnlineTAS: An Online Baseline for Temporal \\ Action Segmentation}

\author{%
  Qing Zhong$^{1,2}$\thanks{Equal Contribution. The work was done when Qing Zhong was an intern in National University of Singapore.} \quad Guodong Ding$^{2*}$ \quad Angela Yao$^{2}$\\
  $^1$University of Adelaide \quad $^2$National University of Singapore\\
  \texttt{qing.zhong@adelaide.edu.au\quad\{dinggd,ayao\}@comp.nus.edu.sg}\\
}
\begin{document}

\maketitle

\begin{abstract} \label{sec:abs}
Temporal context plays a significant role in temporal action segmentation.  
In an offline setting, the context is typically captured by the segmentation network after observing the entire sequence. However, capturing and using such context information in an online setting remains an under-explored problem. 
This work presents the an online framework for temporal action segmentation. At the core of the framework is an adaptive memory designed to accommodate dynamic changes in context over time, alongside a feature augmentation module that enhances the frames with the memory. 
In addition, we propose a post-processing approach to mitigate the severe over-segmentation in the online setting. On three common segmentation benchmarks, our approach achieves state-of-the-art performance.

\end{abstract}

\section{Introduction}
\label{sec:intro}

This work addresses online temporal action segmentation (TAS) of untrimmed videos. Such videos typically feature procedural activities consisting of multiple actions or steps in a loose temporal sequence to achieve a goal~\cite{survey}. For example, ``making coffee'' has actions: `take cup', `pour coffee', `pour water', `pour milk', `pour sugar' and `stir coffee'. 
Standard TAS models~\cite{ms-tcn,yi2021asformer,liu2023diffusion, singhania2023c2f} are offline and segment-only videos of complete procedural activities. 
An \emph{online} TAS model, in contrast, segments only up to the current time point and does not have access to the entire video and, therefore, the entire activity. 

Online TAS faces challenges similar to other online tasks~\cite{lstr,ctvis} in establishing a scalable network that can retain useful information from an ever-increasing volume of data and facilitate effective retrieval when required. 
Additionally, over-segmentation is a common issue for offline TAS, where the segmentation model divides an action into many discontinuous sub-segments, leading to fragmented outputs. This issue is exacerbated in the online setting, as partial data at the onset of an action may lead to erratic predictions and increased over-segmentation.

Most relevant to our task is online temporal action detection (TAD)~\cite{xu2019temporal}. 
Online TAD aims to identify whether an action is taking place and the action category. 
TAD targets datasets like THUMOS~\cite{thumos}, TVSeries~\cite{detec2}, and HACS Segment \cite{hacs}. Among these, 90.8\% videos of THUMOS~\cite{thumos} feature only multiple instances of the same action, while TVSeries~\cite{detec2} comprises diverse yet independent actions (\eg, `open door', `wave' and `write') in one video. 
These actions do not necessarily correlate with one another or impose specific temporal constraints. As such, a direct adaptation of popular online TAD approaches like LSTR~\cite{lstr} and MAT~\cite{unders1} to online segmentation is non-ideal. For instance, these models encode temporal context with a fixed set of tokens, which may limit their capability to handle the 
relations of procedural videos. Furthermore, these models are typically trained to prioritize frame-level accuracy while 
neglecting temporal continuity, which invariably leads to 
over-segmentation. 

To address the online action segmentation task, this work proposes a novel framework centered on a context-aware feature augmentation module and an adaptive memory bank . The memory bank, per-video, tracks short-term and long-term context information. 
The augmentation module uses an attention mechanism to allow frame features to interact with context information from the memory bank and integrate temporal information into standard frame representations. 
Finally, we introduce a post-processing technique for online boundary adjustment that imposes duration and prediction confidence constraints to mitigate over-segmentation.

Summarizing our contributions,
\textbf{1)} We establish an \emph{online} framework for TAS;
\textbf{2)} We propose a feature augmentation module that generates context-aware representations by incorporating an adaptive memory, which accumulates temporal context collectively. The module operates on frame features independently of model architecture, enabling flexible integration;
\textbf{3)} We present a simple post-processing technique for online prediction adjustment, which can effectively mitigate the over-segmentation problem; and 
\textbf{4)} Our framework achieves the state-of-the-art online segmentation performance on three TAS benchmarks.

\section{Related Work}
\label{related_work}

\textbf{Online Action Understanding.}
Many video understanding tasks, such as action detection~\cite{lstr, detec1, detec2, detec3} and video instance segmentation~\cite{minvis, ctvis, dvis}, have been explored in online contexts. For online action detection, videos are classified frame by frame without access to future frames. Specifically, LSTR~\cite{lstr} employs a novel memory mechanism to model long- and short-term temporal dependencies by encoding them as query tokens. Follow-up works feature a segment-based long-term memory compression~\cite{unders1} and fusing short- and long-term histories via attention~\cite{detec3}. %

However, the videos in the datasets commonly used in online TAD contain independent actions~\cite{detec2} or sequences of limited actions~\cite{thumos}, thus lacking temporal relations between the actions. In contrast, TAS deals with untrimmed procedural videos, where such relations are more prominent and may span over long temporal durations.  
There is also a growing trend in online TAD models to use action anticipation as an auxiliary task to enhance action modeling~\cite{detec3, guermal2024joadaa}. In our online segmentation task, we do not assume the availability of such information.

\textbf{Temporal Action Segmentation.} 
In TAS~\cite{survey}, methods vary by their level of supervision, including fully~\cite{ms-tcn, fully1, fully2},  semi-supervised~\cite{ding2022leveraging, semi1}, weakly~\cite{ding2022temporal, weakonline, weak1, weak5, weak2, weak3, weak4}, and unsupervised~\cite{unsuponline, unsup1, unsup2, unsup3, ding2022temporal} setups. 
An emerging direction is to learn TAS incrementally~\cite{ding2024coherent} where procedural activities are learned sequentially. However, all existing works  
are offline, and complete video sequences can be used for inference. In contrast, our approach functions within an online setup. The most related work~\cite{weakonline} investigates online TAS {in a multi-view setup} and leverages the offline model to assist online model learning. Furthermore, it uses the frame-wise multi-view correspondence to generate pseudo-labels for action segments. In contrast, 
we do not assume the availability of multi-view videos nor require assistance from a pre-trained offline model.

\textbf{Post-processing for Action Segmentation.} 
Post-processing methods are either rule-based or leverage graphical modelling.  Rule-based approaches~\cite{otalc, rule1, rule2, 50salads} apply predefined rules to smooth out short-duration predictions that are unlikely given the context. 
Graphical modelling approaches use Conditional Random Fields (CRFs) \cite{crf1,weak3,weak5} to model the relationships and transitions between consecutive actions. 
Online methods need post-processing to alleviate over-segmentation and help locate the action boundaries. 

\section{Online Action Segmentation}
\label{sec:approach}
Previous studies~\cite{ltcontext,sener2020temporal} have demonstrated that the scope of the temporal context significantly influences the performance of (offline) TAS models. This motivates our two lines of inquiry for our online setting: 1) how to consolidate temporal context over an extended period, and 2) how to 
enrich the frame representations with context to benefit the segmentation. %
This work introduces a context-aware feature augmentation module (\Cref{subsec:cfa}) alongside an adaptive context memory (\Cref{subsec:mem}) to tackle the above questions.

\subsection{Preliminaries}
\label{subsec:prelim}
Consider an untrimmed video $v=\{x_t\}_{t=1}^T$ of $T$ frames, where $x_t\in \mathbb{R}^D$ is the per-extracted frame feature at time $t$ and $D$ is the feature dimension. 
An action segmentation model partitions $v$ into $N$ contiguous and non-overlapping temporal segments $\{s_i=(y_i, \ell_i)\}_{i=1}^N$ corresponding to actions $y\in\mathcal{Y}$ present in the video, where $\ell$ indicates the segment length and $\mathcal{Y}$ defines the action space~\cite{survey}. A widely adopted strategy for TAS is to design a segmentation model $\mathcal{M}$ that predicts the action label $y_t$ for each frame $x_t$, akin to a frame-wise classification.  In the offline setting~\cite{ms-tcn,yi2021asformer,liu2023diffusion}, the per-frame prediction $\hat{y}_t$ is based on the entire video sequence. 
The online setting uses only frames up to the current prediction time $t$, without access to future frames. Comparatively:
\begin{equation}\label{eq:offline}
   \hat{y}_t^{\text{offline}}= \argmax_{y\in \mathcal{Y}} p_t(y_t|x_t;x_{1:T}) \quad\text{and} \quad  \hat{y}_t^{\text{online}}=\argmax_{y\in \mathcal{Y}}p_t(y_t|x_t; x_{1:t}).
\end{equation}
where $p_t \in \mathbb{R}^{|\mathcal{Y}|}$ is the estimated action probability for frame $x_t$. 
Most existing offline TAS works~\cite{ms-tcn,yi2021asformer, li2020ms} train the segmentation model $\mathcal{M}$ using a cross-entropy loss ($\mathcal{L}_{\text{cls}}$) for frame-wise classification which is balanced by a smoothing loss ($\mathcal{L}_{\text{sm}}$) that encourages smooth transitions between consecutive frames:
\begin{equation}\label{eq:loss}
    \mathcal{L} = \underbrace{\frac{1}{T} \sum_t -\log(p_t(y_t)) }_{\mathcal{L}_{\text{cls}}}+\lambda\cdot \underbrace{\frac{1}{T|\mathcal{Y}|} \sum_{t,y}\tilde{\Delta}_{t,y}^2}_{\mathcal{L}_{\text{sm}}}, \quad 
\end{equation}
\begin{equation}
   \text{where}\quad  \tilde{\Delta}_{t,y} = \begin{cases}
    \Delta_{t,y}\kern-0.8em &:\! \Delta_{t,y} \le \tau\\
    \tau \kern-0.8em&:\! \text{otherwise}
    \end{cases} \quad \text{and}\quad \Delta_{t,y} = \left|\log p_t(y) - \log p_{t-1}(y)\right|.\nonumber
\end{equation}
In this paper, we opt for the widely recognized convolution-based architecture MS-TCN~\cite{ms-tcn} as our foundational framework. This choice is driven by its relatively lower computational requirements than the attention- or diffusion-based models~\cite{yi2021asformer,liu2023diffusion}. 
A straightforward transition from offline mode to an online mode of the segmentation model $\mathcal{M}$ is to substitute standard convolutions with causal convolutions. Causal and standard convolutions differ in their receptive field in that causal convolutions consider only past and present inputs %
while standard convolutions may incorporate both past and future inputs within a kernel.   
Mathematical details and illustrations of the two are shown in the Appendix.

\subsection{Context-aware Feature Augmentation}
\label{subsec:cfa}
The context-aware feature augmentation (CFA) module generates enhanced clip-wise features through interactions with temporal context captured by an adaptive memory bank. The module operates on a clip-wise basis. During training, video $v$ is split into $K$ non-overlapping clips $\{c_k\}_{k=1}^K$. Each clip has a window size $w$ and is sampled from $v$ with a stride of $\delta = w$, where the final clip 
$c_K$ is padded if $|c_K| < w$. The CFA module integrates the original pre-extracted frame features $c_k=\{x_t\}_{t=(k-1)w+1}^{kw}$ with temporal context to produce a context-enhanced version of representations $\tilde{c}_k=\{\tilde{x}_t\}_{t=(k-1)w+1}^{kw}$. %
Like~\cite{lstr, unders1}, our CFA module is also equipped with a simultaneously updated memory bank $M_k$ as a context resource for feature augmentation. The memory bank is further described in~\Cref{subsec:mem}. 

At each step $k$, 
context is accumulated by feeding $c_k$ through a lightweight GRU~\cite{gru} 
to obtain $c_k^{\text{GRU}}$.  
The GRU is reliable in capturing  
information over long video sequences~\cite{robust}. 
The clip is then passed through a context aggregation block to be augmented. The context aggregation block incorporates the GRU features $c_k^{\text{GRU}}$ with the memory state $M_{k-1}$ from the previous step for $I$ iterations. Concretely, we pass $c_k^{\text{GRU}}$ through a self-attention (SA) block to encourage information exchange with the local clip window.  Additionally, we leverage a Transformer decoder~\cite{vaswani2017attention} to achieve a more effective memory encoding $\tilde{M}_{k-1}^{\text{TD}}$, \ie,
\begin{equation}
\label{eql:self}
   c_k^{\text{SA}} = \text{SelfAttn}(c_k^{\text{GRU}}) \quad \text{and} \quad \tilde{M}_{k-1}^{\text{TD}} = \text{TransDecoder}(M_{k-1}, c_k^{\text{GRU}}, c_k^{\text{GRU}}).
\end{equation}
The outputs from the self-attention module ($c_k^{\text{SA}}$) and the transformer decoder ($c_k^{\text{TD}}$) are then 
combined 
with cross attention (CA) and merged with $c_k^{\text{GRU}}$ to produce the context-augmented features:
\begin{equation}
\label{eql:cross}
    \tilde{c}_k =\text{CrossAttn}(c_k^{\text{SA}}, \Tilde{M}_{k-1}^{\text{TD}},\tilde{M}_{k-1}^{\text{TD}}) + c_{k}^{\text{GRU}}.
\end{equation}
The detailed formulas of $\text{SelfAttn(), TransDecoder(), CrossAttn()}$ are given in the Appendix. An illustration of our CFA module is provided in~\Cref{fig:cfa}.

\definecolor{purple_k}{RGB}{150, 115, 166}
\definecolor{blue_k}{RGB}{108, 142, 191}
\begin{figure}
    \centering
    \begin{overpic}[width=\textwidth]{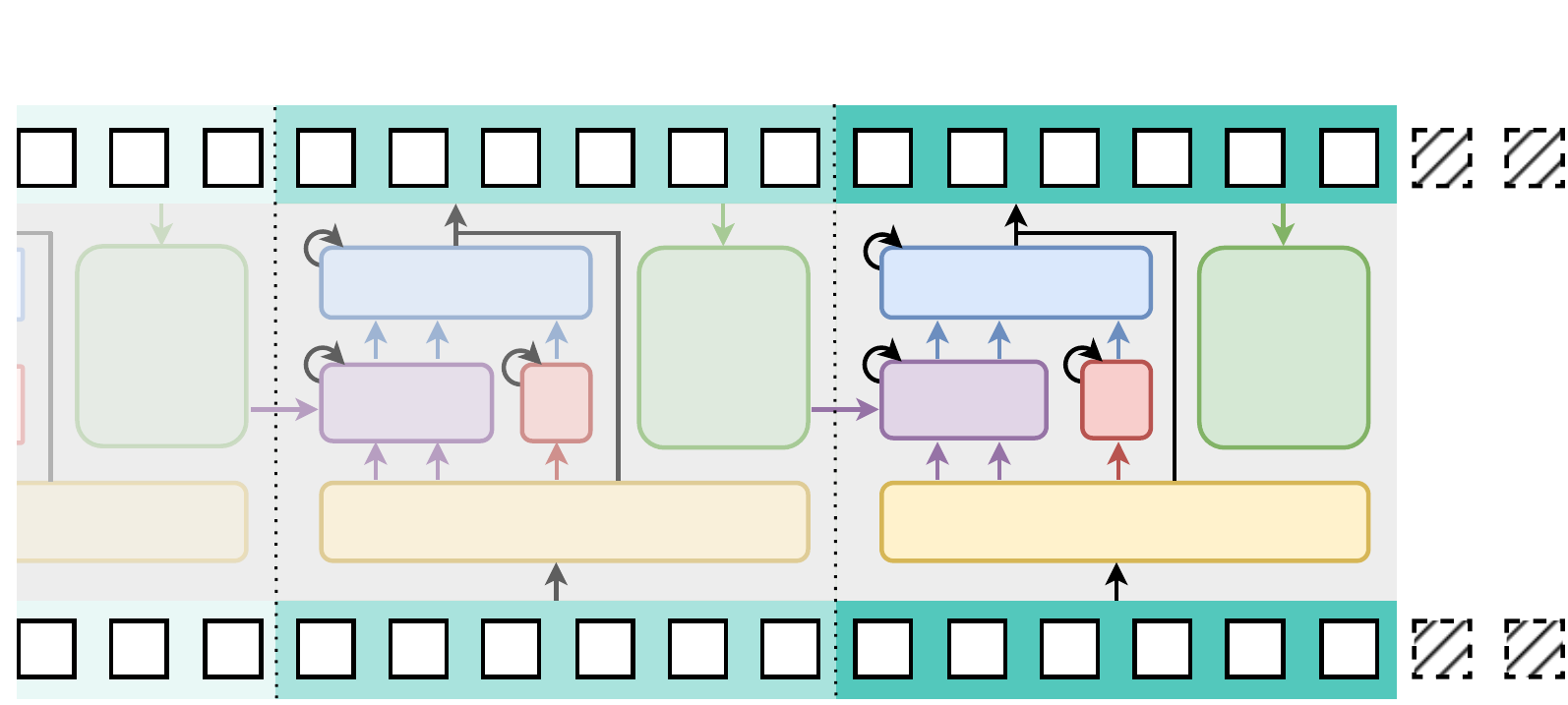}
        \put(1.9,34){\large $\tilde{x}$}
        \put(1.9,2.7){\large $x$}
        \put(85.7,37.5){$t$}
        \put(47.8,37.5){$t\!-\!w$}
        \put(11.5,37.5){$t\!-\!2w$}
        \put(33,39){ $\tilde{c}_{k-1}$}
        \put(69.5,39){ $\tilde{c}_{k}$}
        
        \put(69,11){\small GRU}
        \put(33,11){\small \textcolor{black!60}{GRU}}

        \put(83,29.8){ \textbf{CFA}}
        \put(47.3,29.8){ \textcolor{black!60}{\textbf{CFA}}}
        \put(11.8,29.8){ \textcolor{black!30}{\textbf{CFA}}}
        \put(69.8,18.5){\small SA}
        \put(34,18.5){\small \textcolor{black!60}{SA}}
        \put(63,26.2){\small CA}
        \put(19.2,30.5){\tiny \textcolor{black!60}{$I$}}
        \put(32,23){\tiny \textcolor{black!60}{$I$}}
        \put(19.2,23){\tiny \textcolor{black!60}{$I$}}
        \put(57.6,22.8){\tiny \textcolor{blue_k}{$K$}}
        \put(61.7,22.8){\tiny \textcolor{blue_k}{$V$}}
        \put(72.4,22.8){\tiny \textcolor{blue_k}{$Q$}}
        \put(27.5,26.2){\small \textcolor{black!60}{CA}}
        \put(21.9,22.8){\tiny \textcolor{blue_k!60}{$K$}}
        \put(26,22.8){\tiny \textcolor{blue_k!60}{$V$}}
        \put(36.5,22.8){\tiny \textcolor{blue_k!60}{$Q$}}
        
        \put(57.8,19.7){\small Trans.}
        \put(57.8,17.5){\small Decoder}
        \put(57.6,15){\tiny \textcolor{purple_k}{$K$}}
        \put(61.7,15){\tiny \textcolor{purple_k}{$V$}}
        \put(54.2,17){\tiny \textcolor{purple_k}{$Q$}}
        \put(55.2,23){\tiny {$I$}}
        \put(68,23){\tiny {$I$}}
        \put(55.2,30.5){\tiny {$I$}}
        \put(22.2,19.7){\small \textcolor{black!60}{Trans.}}
        \put(22.2,17.5){\small \textcolor{black!60}{Decoder}}
        \put(21.9,15){\tiny \textcolor{purple_k!60}{$K$}}
        \put(26,15){\tiny \textcolor{purple_k!60}{$V$}}
        \put(18.5,17){\tiny \textcolor{purple_k!60}{$Q$}}
        \put(78,25.2){\small Adaptive }
        \put(78,23){\small Memory }
        \put(78,20.8){\small Bank }
        \put(78,18.2){\small $M_{k}$}

        \put(93.5,19){. . .}

        \put(33,-1){ $c_{k-1}$}
        \put(69.5,-1){ $c_{k}$}
        \put(42,25.2){\small \textcolor{black!60}{Adaptive }}
        \put(42,23){\small \textcolor{black!60}{Memory }}
        \put(42,20.8){\small \textcolor{black!60}{Bank }}
        \put(42,18.2){\small \textcolor{black!60}{$M_{k-1}$ }}
        
         \put(6.5,25.2){\small \textcolor{black!30}{Adaptive }}
        \put(6.5,23){\small \textcolor{black!30}{Memory }}
        \put(6.5,20.8){\small \textcolor{black!30}{Bank }}
        \put(6.5,18.2){\small \textcolor{black!30}{$M_{k-2}$ }}
        \put(2,11){\small \textcolor{black!30}{GRU}}
    \end{overpic}
    \caption{Context-aware Feature Augmentation (CFA) module. CFA takes as input a video clip $c_{k}$ of length $w$, augments it with temporal information captured in an adaptive memory bank $M_k$, and outputs an enhanced clip feature $\tilde{c}_{k}$. $I$ is the number of iterations of SA, TransDecoder, and CA.}
    \label{fig:cfa}
\end{figure}
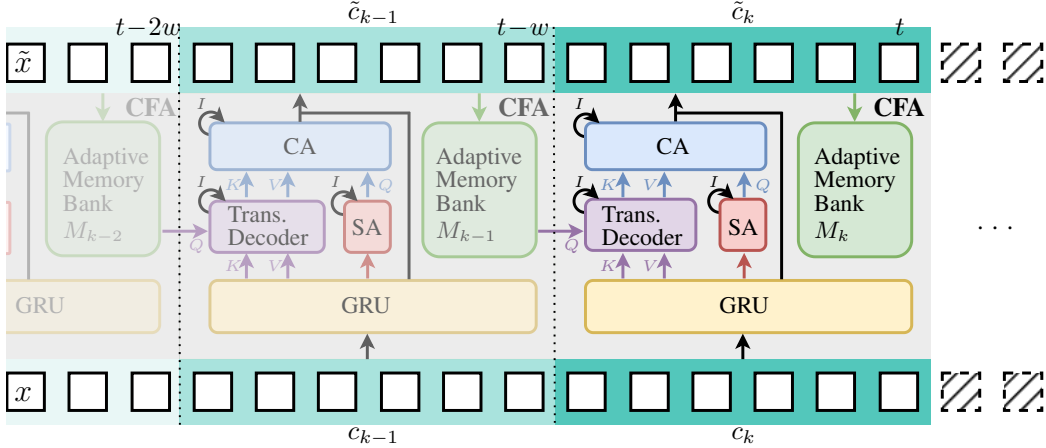

\subsection{Adaptive Memory Bank}
\label{subsec:mem}

In a similar spirit with~\cite{lstr}, our memory is designed to account for both short- and long-term context, \ie, $M = [M^{\text{long}},M^{\text{short}}]$. Short-term memory helps capture the local action dynamics while long-term memory 
retains information across extended durations important for TAS~\cite{ltcontext,sener2020temporal}. 

\textbf{Short Memory} $M^\text{short}$. Given that our enhancement module works on a per-clip basis with temporal stride $w$, we directly regard the enhanced feature $\tilde{c}_{k-1}$ from the last clip as the short-term memory, \ie, $M_k^{\text{short}} = \tilde{c}_{k-1} \in \mathbb{R}^{Dw}$. 

\begin{wrapfigure}{R}{0.45\textwidth}
\vspace{-2em}
\begin{minipage}{0.45\textwidth}
\begin{algorithm}[H]
\caption{Adaptive Memory Update}\label{alg:amu}
\begin{algorithmic}[1]
\Require $\{c_k\}_{k=1}^K$ , $w$ 
\State Initialize $M_0^{\text{short}} \gets c_1, M_0^{\text{long}} \gets \varnothing$
\For{$k \in [1 ... K]$}
        \State $\tilde{c}_k = \text{CFA}(c_k, M_{k-1})$ %
        \State $m_k =\text{Conv1D}(\tilde{c}_k)$
	\If{$\text{len}(M_{k-1}^{\text{long}}) \leq \frac{2}{3}w$}
        \State $M_k^{\text{long}} = \text{concat}(M_{k-1}^{\text{long}}, m_k)$
        \Else
            \State $M_k^{\text{long}} = \text{concat}(M_{k-1}^{\text{long}}[1:], m_k)$
        \EndIf
        \State $M_k^{\text{short}} = \tilde{c}_{k-1}[\text{len}(M_k^{\text{long}}):]$
        \State $M_k = [M_k^{\text{long}}, M_k^{\text{short}}]$
\EndFor
\end{algorithmic}
\end{algorithm}
\end{minipage}\vspace{-1em}
\end{wrapfigure}
\textbf{Long Memory} $M^{\text{long}}$. We update our long-term memory with information from processed clips. Specifically, we apply a convolutional layer on top of our context-enhanced representation $\tilde{c}_k$, where $\tilde{c}_k\in \mathbb{R}^{Dw}$, to collapse the temporal dimension and yield 
a memory token $m_k=\text{Conv1D}(\tilde{c}_k) \in \mathbb{R}^{D}$. 
This memory token is then appended to the current long-term memory.

\textbf{Adaptive Memory Update.} The memory is updated whenever a new clip is processed. 
In practice, we constrain the total footprint of both short- and long-term memory to match the size of the processing clip, \ie, $M\in\mathbb{R}^{Dw}$. At the beginning of each sequence, the memory bank is initialized with short-term information only, \ie, $M\!=\!M_0^{\text{short}}\!=\!c_1$ and $M_0^{\text{long}}\!=\!\varnothing$. 
As more clips are processed, we gradually increase the budget to accommodate longer-term information. However, in anticipation of $M^{\text{long}}$ draining the entire budget in prolonged sequences, we %
cap its utilization at a maximum of two-thirds of the total budget. In instances where this threshold is exceeded, the earliest token is discarded, \ie,:
\begin{equation}\label{eq:long}
    M_k^{\text{long}} = \begin{cases}
        \text{concat}(M_{k-1}^{\text{long}}, m_k) &:\text{len}(M_{k-1}^{\text{long}}) \le \frac{2}{3} w \\[0.5em]
        \text{concat}(M_{k-1}^{\text{long}}[1\!:], m_k) &: \text{otherwise}
    \end{cases}.
\end{equation}

The remaining budget is allocated for the short-term information accordingly:
\begin{equation}\label{eq:short}
     M_k^{\text{short}} = \tilde{c}_{k-1} [\text{len}(M_k^{\text{long}})\!:]
\end{equation}
As video progresses, our feature augmentation module (\Cref{subsec:cfa}) receives more longer-term context while emphasizing only shorter and more relevant short-term information. 
This adaptive approach enables the context memory to flexibly shift its attention between short and long-term information as the video progresses. \Cref{alg:amu} summarizes the update mechanism.

\textbf{Discussion.} Our module integrates context memory on top of a GRU layer. While the GRU captures temporal dependencies, it 
may struggle, especially in thousands-frame long sequences common in TAS. The context memory supplements the GRU by allowing selective access and updates, thereby enabling the retrieval and manipulation of long-term information. Such a design is supported by our empirical study that the explicit memory can extend the capacity of the GRU's internal state. Supporting ablations are found in~\Cref{subsec:abl}. 

\begin{figure}[t]
    \centering
    \begin{overpic}[width=1\textwidth]{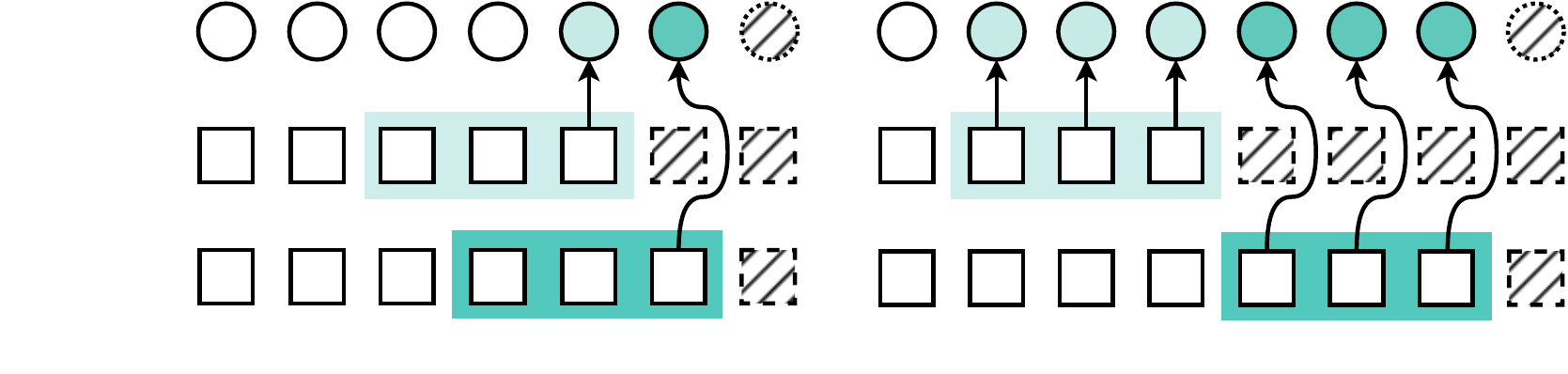}
        \put(15,-2){(a) Online Inference ($\delta\!=\!1$)}
        \put(60,-2){(b) Semi-online Inference ($\delta\!=\!w$) }
        \put(36.5,9.8){\small $x_k$}
        \put(73.5,9.8){\small $x_{kw}$}
        \put(41.5,1.5){\small $x_{k+1}$}
        \put(89.7,1.5){\small $x_{(k+1)w}$}
        \put(1,21){\small Output $\hat{y}:$}
        \put(1,13){\small Step $k:$}
        \put(1,5){\small Step $k\!+\!1:$}
    \end{overpic}
    \vspace{0.3em}
    \caption{Two inference types. a) Online inference samples clips with stride $\delta=1$ and  only preserves the last frame prediction, while b) Semi-online inference samples non-overlapping clips with stride $\delta=w$ and all predictions are preserved.}
    \label{fig:inference}
\end{figure}

\begin{wrapfigure}{R}{0.45\textwidth}
\vspace{-2.2em}
\begin{minipage}{0.45\textwidth}
\begin{algorithm}[H]
\caption{Post-processing for Online TAS}\label{alg:post}
\begin{algorithmic}[1]
\State Compute $\ell_{\text{min}} = \sigma \times T_\text{max}$
\State Initialize $\ell = 0$
\For{each frame $t$}
    \If {$q_t < \theta$ and $\ell < \ell_{\text{min}}$}
        \State $\hat{y}^*_t = \hat{y}^*_{t-1}$
        \State $\ell = \ell + 1$
    \Else
        \State $\hat{y}^*_t = \hat{y}_t$
        \State $\ell = 0$
    \EndIf
\EndFor
\end{algorithmic}
\end{algorithm}
\end{minipage}
\vspace{-0.2em}
\end{wrapfigure}

\subsection{Training and Inference}
\label{subsec:inference}
\textbf{Training.} Our final online segmentation  model is constructed by combining our CFA module with a single-stage TCN~\cite{ms-tcn} with causal convolutions. Specifically, TCN takes as input the enhanced representations $\tilde{c}_k$ and maps them to the same labels for $c_k$. We train the framework end-to-end with the loss function formulated in~\Cref{eq:loss}, but on a clip basis with $T$ replaced by $w$:
\begin{equation}
    \mathcal{L}^{\text{clip}} = \frac{1}{w} \sum_t -\log(p_t(y_t)) +\lambda\cdot \sum_{t,y}\tilde{\Delta}_{t,y}^2,
\end{equation}
we set $\lambda=0.15$ following~\cite{ms-tcn}. Like~\cite{lstr}, training on a clip basis provides better efficiency.

\textbf{Inference.}
We present two distinct inference approaches by manipulating the clip sampling stride parameter $\delta$. 
The first mode of inference, referred to as \emph{online}, is characterized by setting $\delta=1$. In such a setting, a video clip of $w$ frames is processed, with emphasis placed solely on the prediction derived from the final frame and rest are discarded. 
This facilitates the scenario when frame-by-frame prediction is preferred. 
The alternate mode of inference, termed \emph{semi-online}, adheres to the training regime by setting $\delta=w$. In this mode, video clips are processed, and the dense predictions generated across all $w$ frames are preserved as final output. 
An illustration of two modes of inference is provided in~\Cref{fig:inference}. %

\subsection{Post-processing}
\label{subsec:postproc}

Our intuition is that a valid action segment should not fall below a minimum length threshold unless there is high confidence in the prediction to justify a change in the action class. Specifically, we consider the maximum softmax probability of a prediction as its confidence measure, denoted as $q_t = \max(p_t)$, for frame $x_t$, as $q_t$ to some extent indicates its reliability. A prediction is considered ``unreliable'' if its confidence measure scores below a certain threshold $\theta$, \ie, $q_t<\theta$. For the frame with ``unreliable'' prediction, we disregard its current prediction and assign the action label of its proceeding frame $\hat{y}_{t-1}^*$, when the previous action segment is shorter than the minimum length. Otherwise, we retain the original prediction. We set the length threshold $\ell_{\text{min}}=\sigma\times T_\text{max}$ with $\sigma \in (0,1)$, in proportion to to the longest video duration $T_\text{max}$ in training set.  
 
Our post-processing mitigates the over-segmentation by adjusting action boundaries according to network predictions and action length, which is very efficient compared to~\cite{unsup2,unsup3} that calculates frame similarities. The procedure for post-processing is illustrated in~\Cref{alg:post}. %

\section{Experiments}\label{sec:exp}
\textbf{Datasets}: We evaluate our model on three common TAS datasets. 
\textbf{Breakfast}~\cite{breakfast} comprises in total 1,712 videos performing ten different activities with 48 actions. On average, each video contains six action instances.
\textbf{50Salads}~\cite{50salads} has 50 videos with 17 action classes. \textbf{GTEA}~\cite{gtea} contains 28 videos of seven kitchen activities composing 11 different actions. We use common I3D features~\cite{i3d} as input. 

\textbf{Evaluation Metrics}: Standard evaluation metrics for TAS are reported for our online setting, which includes frame-wise accuracy (Acc), segmental edit score
(Edit), and segmental F1 scores with varying overlap thresholds 10\%, 25\%, and 50\%.

\textbf{Implementation Details.} 
In CFA, we stack 2 Transformer decoder layer with 8 heads, 2 Swin~\cite{swin} self- and cross attention with 4 heads. 
We use a single-stage TCN as segmentation backbone and sample non-overlapping clips \ie, $\delta=w$ for efficiency.  
We train the model end-to-end with a learning rate of $5e^{-4}$ of total 50 epochs. 
Detailed hyperparameter settings can be found in Appendix. 
\begin{table}[t]
\centering
\scalebox{0.9}{
\begin{tabular}{lccccccccccccccccc}
\toprule
\multirow{3}{*}{} & \multirow{3}{*}{{p.p.}} & \multicolumn{5}{c}{{GTEA~\cite{gtea}}} & \multicolumn{5}{c}{{50Salads~\cite{50salads}}}&\multicolumn{5}{c}{{Breakfast~\cite{breakfast}}} \\ 
\cmidrule(lr){3-7} \cmidrule(lr){8-12}\cmidrule(lr){13-17}
& & Acc & Edit & \multicolumn{3}{c}{F1 @ \{10, 25, 50\}} & Acc & Edit & \multicolumn{3}{c}{F1 @ \{10, 25, 50\}} & Acc & Edit & \multicolumn{3}{c}{F1 @ \{10, 25, 50\}} \\ 
\midrule
\multicolumn{17}{c}{Online}\\
\midrule
& - &74.4 & 66.6 & 73.9 & 70.3 & 57.2 & 75.2 & 19.6 & 26.8 & 24.4 & 19.6&55.3 & 18.7 & 15.1 & 11.7 & 8.3\\\rowcolor{gray!30}
\multirow{-2}{*}{\rotatebox[origin=c]{90}{TCN}}& \checkmark &72.1 & \textbf{71.9} & \textbf{79.2} & \textbf{77.4} & 64.1 & 75.1 & 68.5 & 74.1 & 70.6 & 60.4 & 52.3 & 54.7 & 52.0 & 43.2 & 29.8\\
\cmidrule(lr){1-2}\cmidrule(lr){3-7} \cmidrule(lr){8-12}\cmidrule(lr){13-17}
& - & \textbf{76.2} & 63.5 & 72.6 & 68.3 & 58.8 & \textbf{80.9} & 28.8 & 36.1 & 31.0 & 23.3 &\textbf{56.7}&19.3&16.8&13.9&9.3\\ \rowcolor{gray!30}
 \multirow{-2}{*}{\rotatebox[origin=c]{90}{Ours}}& \checkmark & 74.4 & 70.3 & 78.5 & 76.4 & \textbf{67.7} & 77.7 & \textbf{71.5} & \textbf{77.7} & \textbf{74.6} & \textbf{64.1} &52.9&\textbf{55.7}&\textbf{54.8}&\textbf{45.8}&\textbf{30.5}\\ %
\midrule
\multicolumn{17}{c}{Semi-online}\\
\midrule

& - &
75.8 & 66.8 &74.3 &71.5 &60.3 &79.1 & 29.0 & 38.5 & 35.5 & 28.3&55.7 & 18.6 & 15.4 & 12.7 & 9.0\\\rowcolor{gray!30}
\multirow{-2}{*}{\rotatebox[origin=c]{90}{TCN}}&\checkmark & 73.5 &75.4 & 80.3 & 76.9 & 66.6 & 76.7 & 69.2 & 73.1 & 70.5 & 62.8 &52.5 & 54.0 & 53.1 & 44.5 & 29.6\\\cmidrule(lr){1-2}\cmidrule(lr){3-7} \cmidrule(lr){8-12}\cmidrule(lr){13-17}

& - & \textbf{77.1} & 68.1 & 76.7 & 73.5 & 63.9 & \textbf{82.4} & 32.8 & 43.0 & 41.1 & 34.7 &\textbf{57.4} & 19.6 & 17.8 & 14.8 & 10.1 \\ \rowcolor{gray!30}
\multirow{-2}{*}{\rotatebox[origin=c]{90}{Ours}} & \checkmark & 76.0 & \textbf{79.7} & \textbf{84.9} & \textbf{81.4} & \textbf{69.2} & 79.4 & \textbf{75.0} & \textbf{82.5} & \textbf{80.2} & \textbf{68.0}&53.8 & \textbf{57.5}& \textbf{56.4} & \textbf{47.3} & \textbf{31.4} \\ 
\bottomrule
\end{tabular}}
\caption{Performance of our approach on three TAS benchmarks under two inference mode, \ie, online and semi-online. Post-processing is indicated by p.p..}
\label{tab:mem_inf_eval}
\end{table}
\subsection{Experiment Results}
\label{subsec:abl}

\textbf{Effectiveness.} %
We report the overall performance for \textit{online} and \textit{semi-online} inference (see~\Cref{subsec:inference}) in in~\Cref{tab:mem_inf_eval}. Our baseline is a single-stage causal TCN and we build our framework on top of it. 
Across all three datasets, the integration of our CFA module leads to a consistent boost in the segmentation performance. Specifically, our approach gained 5.7\% (75.2\% vs. 80.9\%) in Acc and 9.2\% (19.6\% vs. 28.8\%) in Edit on 50Salads. While the improvements on other datasets are not as significant, they still show effectiveness, with a margin of about 2\%. 
Generally, semi-online inference achieves better performance over online across all metrics. Such improvement is likely because clip-wise prediction better preserves the local temporal continuity of labels compared to step-by-step single frame prediction. 

Comparing across the metrics, segmental scores appear to be significantly low. On breakfast~\cite{breakfast} with our online inference, a frame-wise accuracy of 56.7\% only corresponds to a 9.3\% F1 score with 50\% IoU. Such score indicates a severe over-segmentation issue and necessitates an effective post-processing. However, a significant performance increase is observed on Edit and F1 scores after our proposed pose-processing. For example, the same F1 score increases to 30.5\%, tripling its original value. Although post-processing could lead to a slight decrease in accuracy, it demonstrates great effectiveness in mitigating the over-segmentation problem.

\textbf{Ablation study.} %
\Cref{tab:component_impact} evaluates the components in our CFA module. The first row is our single-layer causal TCN baseline %
with strong frame-wise accuracy but poor %
segmental metrics. 
The GRU %
boosts segment metrics (7-11\%) over the baseline, showing its %
ability to accumulate context information. While CFA using the current clip as pseudo memory predictably leads to a performance drop (5\%) compared to GRU due to lack of any context information.  Combining either GRU or our adaptive memory with our CFA achieves very close performance (rows 4 and 5), highlighting the importance of the context information for TAS. The complete model yields the best performance and boosts Acc by 7\% and average segmental scores by 15.3\%. This validates the complementary effect of GRU's internal state and our explicit memory design. %

\begin{table}[t]
    \begin{minipage}{0.5\textwidth}
        \centering
        \scalebox{1}{
\begin{tabular}{cccccccc}
\toprule
{GRU} & {CFA} & {Mem.} & {Acc} & {Edit} & \multicolumn{3}{c}{{F1 @ \{10, 25, 50\}}} \\
\cmidrule(lr){1-3}\cmidrule(lr){4-8}
- & - & - & 75.2 & 19.6 & 26.8 & 24.4 & 19.6 \\
\checkmark & - & - & 78.1 & 27.1 & 37.9 & 34.7 & 26.7 \\
- & \checkmark & - & 76.2 & 22.3 & 30.1 & 27.0 & 21.9 \\
\checkmark & \checkmark & - &79.1 & 29.0 & 38.5 & 35.5 & 28.3 \\
- & \checkmark & \checkmark & 78.9 & 29.2 & 38.7 & 35.1 & 28.8 \\\rowcolor{gray!30}
\checkmark & \checkmark & \checkmark & \textbf{82.4} & \textbf{32.8 }& \textbf{43.0} & \textbf{41.1} & \textbf{34.7} \\
\bottomrule
\end{tabular}}
\caption{Ablation study of module components on 50Salads~\cite{50salads}.}
\label{tab:component_impact}
    \end{minipage}\hfill
    \begin{minipage}{0.43\textwidth}
        \centering
        \scalebox{1}{
\begin{tabular}{cccccc}
\toprule
{$I$} & {Acc} & {Edit} & \multicolumn{3}{c}{{F1 @ \{10, 25, 50\}}} \\ \cmidrule(lr){1-1}\cmidrule(lr){2-6}
1 & 79.1 & 29.0 & 38.5 & 35.5 & 28.3 \\ \rowcolor{gray!30}
2 & \textbf{79.6 }& 30.7 &\textbf{ 40.7} & \textbf{38.2} & \textbf{31.4} \\ 
3 & 79.5 & 28.5 & 37.2 & 36.1 & 29.0 \\ 
4 & 79.1 & 29.2 & 39.1 & 36.3 & 30.5 \\ 
5 & 79.2 & \textbf{30.8} & 39.1 & 37.7 & 30.6 \\ 
\bottomrule
\end{tabular}}
\caption{Effect of interactions $I$.}
\label{tab:swin_layers}
\end{minipage}
\end{table}

\textbf{Number of layers in CFA.} \Cref{tab:swin_layers} explores the interaction iterations $I$ in CFA. The results indicate that the performance is not significantly affected by the number of iterations. In practice, we set the number of iterations to 2, as it achieves a good balance between performance and efficiency.  %

\begin{table}[h]
\centering
\begin{minipage}{0.4\textwidth}
        \centering
    \begin{tabular}{cccc}
    \toprule
    {$M^\text{short}$} & {$M^\text{long}$} & {Acc} & {Seg.} \\ \cmidrule(lr){1-2} \cmidrule(lr){3-4}
    \checkmark & - & 80.3 & 36.7 \\ 
    - & \checkmark & 80.4 & 36.4 \\ \rowcolor{gray!30}
    \checkmark & \checkmark & \textbf{82.4} & \textbf{37.9} \\ 
    \bottomrule
    \end{tabular}
    \caption{Effect of memory composition.}
    \label{tab:mem_type}
\end{minipage}%
\hspace{1.2em}
\begin{minipage}{0.5\textwidth}
    \centering
    \begin{tabular}{cccccc}
    \toprule
    $w$ / $\text{len}(M)$ & {16} & {32} & {64} & {128} & {256} \\ \cmidrule(lr){1-1}\cmidrule(lr){2-6}
    {Acc} & 79.8 & 81.4 & 81.6 & \cellcolor{gray!30}{\textbf{82.4}} & 80.7 \\ 
    {Seg.} & 35.1 & 36.5 & 37.6 & \cellcolor{gray!30}{37.9} & \textbf{38.2} \\ 
    \bottomrule
    \end{tabular}
    \caption{Effect of clip size and memory length. ``Seg.'' indicates the mean of Edit and F1 scores.}
    \label{tab:slmem}
\end{minipage}
\end{table}

\textbf{Memory composition} ($M^{\text{short}}$/$M^{\text{long}}$). We assess the impact of memory types and present results in~\Cref{tab:mem_type}. It shows comparable performances for each memory type when considered individually. However, the combination of both yields a 2\% improvement in Acc and 1.5\% for averaged segmental metric, suggesting the significance of incorporating diverse memory for TAS.

\textbf{Clip size $w$ / Memory size} $\text{len}(M)$. In our implementation, we set clip size and memory size to be equal and we report its influence on performance in~\Cref{tab:slmem}. It shows a larger clip size leads to better segmental results; this is because temporal continuity can be better modeled with longer clips for learning. However, the information of short actions could be diluted when compressed to form the memory token if the window size is too large. For memory sizes of 16, 32, and 64, the earliest memory are discarded as the average length of 50Salads is $\approx$5.8k frames. With the size reducing, the performance gradually drops and reaches the lowest of 79.8\% in Acc compared to the peak of 82.4\%. Note that with the memory size set to 16, our approach only retains long-term information from up to 192 frames, 30$\times$ less than the average video length.

\textbf{Post-processing hyperparameters.} Two hyperparameters are defined in our post-processing: confidence threshold $\theta$ and the minimum segment length $\ell_{\text{min}}$. 
We vary its scaling factor $\sigma$ to assess $\ell_{\text{min}}$. 
In~\Cref{tab:theta}, increasing $\theta$ greatly enhances the segment results, with a 18.1\% increase observed when $\theta = 0.7$. Although the accuracy tends to decrease as $\theta$ becomes larger, the drop is not as substantial (3.1\%) compared to the improvements in segmental results. 
While \Cref{tab:lmin} shows the segmental performance stops increasing and stays stable when $\sigma>\frac{1}{16}$ with a fixed confidence score $\theta = 0.9$. In conclusion, employing a higher confidence threshold can help better mitigate the over-segmentation because it makes more sense to prioritize preserving the continuity of a segment that includes frames with highly confident predictions given a fixed length budget.

\begin{table}[htb]
    \begin{minipage}{0.5\textwidth}
        \centering
\begin{tabular}{lccccccc}
\toprule
$\theta$ & {0.3} & {0.4} & {0.5} & {0.6} & {0.7} & {0.8} & {0.9} \\ \cmidrule(lr){1-1}\cmidrule(lr){2-8}
Acc & 82.5 & \textbf{82.6} & 82.6 & 82.6 & 81.2 & 81.1 & \cellcolor{gray!30}{79.4 }\\ 
{Seg.} & 50.9 & 51.8 & 51.9 & 51.9 & 69.0 & 73.3 & \cellcolor{gray!30}{\textbf{76.4}} \\ 
\bottomrule
\end{tabular}
\caption{Effect of confidence threshold $\theta$ ($\sigma\!=\!\frac{1}{16}$). }
\label{tab:theta}
\end{minipage}\hspace{1.5em}
    \begin{minipage}{0.44\textwidth}
        \centering
\begin{tabular}{lccccc}
\toprule
$\sigma$& $\frac{1}{64}$ & $\frac{1}{32}$& $\frac{1}{16}$ & $\frac{1}{8}$ & $\frac{1}{4}$ \\ \cmidrule(lr){1-1}\cmidrule(lr){2-6}
Acc & \textbf{79.6} & 79.4 &  \cellcolor{gray!30}{79.4} & 79.4 & 79.4 \\ 
Seg. & 70.9 & 74.7 & \cellcolor{gray!30}{\textbf{76.4}} & 76.4 & 76.4 \\ 
\bottomrule
\end{tabular}
\caption{Effect of minimum length factor $\sigma$ with $\theta = 0.9$. }
\label{tab:lmin}
    \end{minipage}
\end{table}

\begin{table}[h]
\centering
\begin{tabular}{llcccccccccc}
\toprule
\multirow{3}{*}{{}} & \multirow{3}{*}{{Method}} & \multicolumn{5}{c}{{GTEA~\cite{gtea}}} & \multicolumn{5}{c}{{50Salads~\cite{50salads}}} \\ 
\cmidrule(lr){3-7} \cmidrule(lr){8-12}
& & Acc & Edit & \multicolumn{3}{c}{F1 @ \{10, 25, 50\}} & Acc & Edit & \multicolumn{3}{c}{F1 @ \{10, 25, 50\}} \\ 
\cmidrule(lr){1-2}\cmidrule(lr){3-7} \cmidrule(lr){8-12}
\multirow{4}{*}{\rotatebox[origin=c]{90}{offline}} & MS-TCN~\cite{ms-tcn} & 78.7 & 84.0 & 88.3 & 86.6 & 72.8 & 81.2 & 65.8 & 72.8 & 70.4 & 61.7 \\ 
& MS-TCN + p.p. & 78.7 & 85.2 & 89.6 & 88.3 & 73.3 & 80.4 & 74.1 & 82.0 & 79.2 & 70.2 \\ 
& ASFormer~\cite{yi2021asformer} & 79.7 & 84.6 & 90.1 & 88.8 & 79.2 & 85.6 & 79.6 & 85.1 & 83.4 & 76.0 \\ 
& DiffAct~\cite{liu2023diffusion} & 82.2 & 89.6 & 92.5 & 91.5 & 84.7 & 87.4 & 88.9 & 90.1 & 89.2 & 83.7\\
\cmidrule(lr){1-2}\cmidrule(lr){3-7} \cmidrule(lr){8-12}
\multirow{6}{*}{\rotatebox[origin=c]{90}{online}} & LSTR~\cite{lstr} & 63.7 & 33.2 & 41.5 & 37.7 & 25.0 & 60.5 & 5.0 & 8.2 & 6.6 & 4.1 \\ 
& Causal TCN & 74.4 & 66.6 & 73.9 & 70.3 & 57.2 & 75.2 & 19.6 & 26.8 & 24.4 & 19.6 \\ 
& Ours$^\text{online}$ & 75.8 & 66.8 & 74.3 & 71.5 & 60.3 & 79.1 & 29.0 & 38.5 & 35.5 & 28.3 \\ 
& Ours$^\text{online}$ + p.p. & 73.5 & 75.4 & 80.3 & 76.9 & 66.6 & 76.7 & 69.2 & 73.1 & 70.5 & 62.8 \\ 
& Ours$^\text{semi}$ & \textbf{77.1} & 68.1 & 76.7 & 73.5 & 63.9 & \textbf{82.4} & 32.8 & 43.0 & 41.1 & 34.7 \\ \rowcolor{gray!30}
& Ours$^\text{semi}$ + p.p. & 76.0 & \textbf{79.7} & \textbf{84.9} & \textbf{81.4} & \textbf{69.2} & 79.4 & \textbf{75.0} & \textbf{82.5} & \textbf{80.2} & \textbf{68.0} \\ 
\bottomrule
\end{tabular}
\caption{Comparison with the state-of-the-art methods on GTEA and 50Salads.}
\label{tab:method_comparison}
\end{table}

\subsection{Comparison with State-of-the-Art Methods} \label{subsec:sota}
\Cref{tab:method_comparison,tab:method_comparison_break} compare our approach against the state-of-the-art TAS approaches on all three benchmarks. Due to the absence of dedicated online TAS methods, we benchmark against the online TAD approach LSTR~\cite{lstr}. 
We train LSTR on TAS datasets using the official code implementation\footnote{\url{https://github.com/amazon-science/long-short-term-transformer}}. To ensure a fair comparison, we configure their working (short-term) memory to be the same as ours ($w$). Additionally, we adjust its long memory %
accordingly to provide access to the entire past sequence. As evident from~\Cref{tab:method_comparison,tab:method_comparison_break}, LSTR~\cite{lstr} consistently achieves relatively low performance, particularly with Edit scores of 5.0\% and 4.9\% on 50Salads and Breakfast datasets, respectively. This suggests severe over-segmentation in their predictions. Moreover, these performances are %
inferior even to those of our baseline model (casual TCN), indicating that a direct adoption of online detection models for the segmentation task is not ideal.

Amongst all datasets, Breakfast %
is the most challenging, with a significant performance gap between offline and online models, particularly on segmental metrics. Notably, the F1@50 score on Breakfast experiences a drastic drop of 4/5, from 47.5\% to 8.3\%, highlighting the difficulty of the online segmentation task with videos that are more complex. 
Nonetheless, we %
still achieve a %
modest absolute performance improvement of 2\%. 
Furthermore, our post-processing technique, significantly boosts segmental performance, nearly tripling the original performance, albeit with a slight decrease in Acc. %
This underscores the effectiveness of our post-processing technique in mitigating the over-segmentation. 
MV-TAS~\cite{weakonline} tackles online segmentation but under a multi-view setting. It leverages multi-view information and an offline model as a reference for online segmentation. Despite this, even our baseline model, depicted in the third-to-last row of~\Cref{tab:method_comparison_break}, showcases a notable performance improvement (55.3\% vs. 41.6\%) over MV-TAS~\cite{weakonline}.  This considerable margin emphasizes the competitiveness of our baseline model. 

When compared to offline models, our semi-online inference with post-processing manages to surpass the offline model MS-TCN~\cite{ms-tcn} on 50Salads dataset across the segmental metrics and reaches around 90\% of the accuracy of the best-performing DiffAct~\cite{liu2023diffusion}. %
On Breakfast, our approach lags behind the offline model %
in both frame-wise accuracy and segmental metrics.

\begin{figure}[htb]
\centering

\begin{overpic}[width=0.8\textwidth]{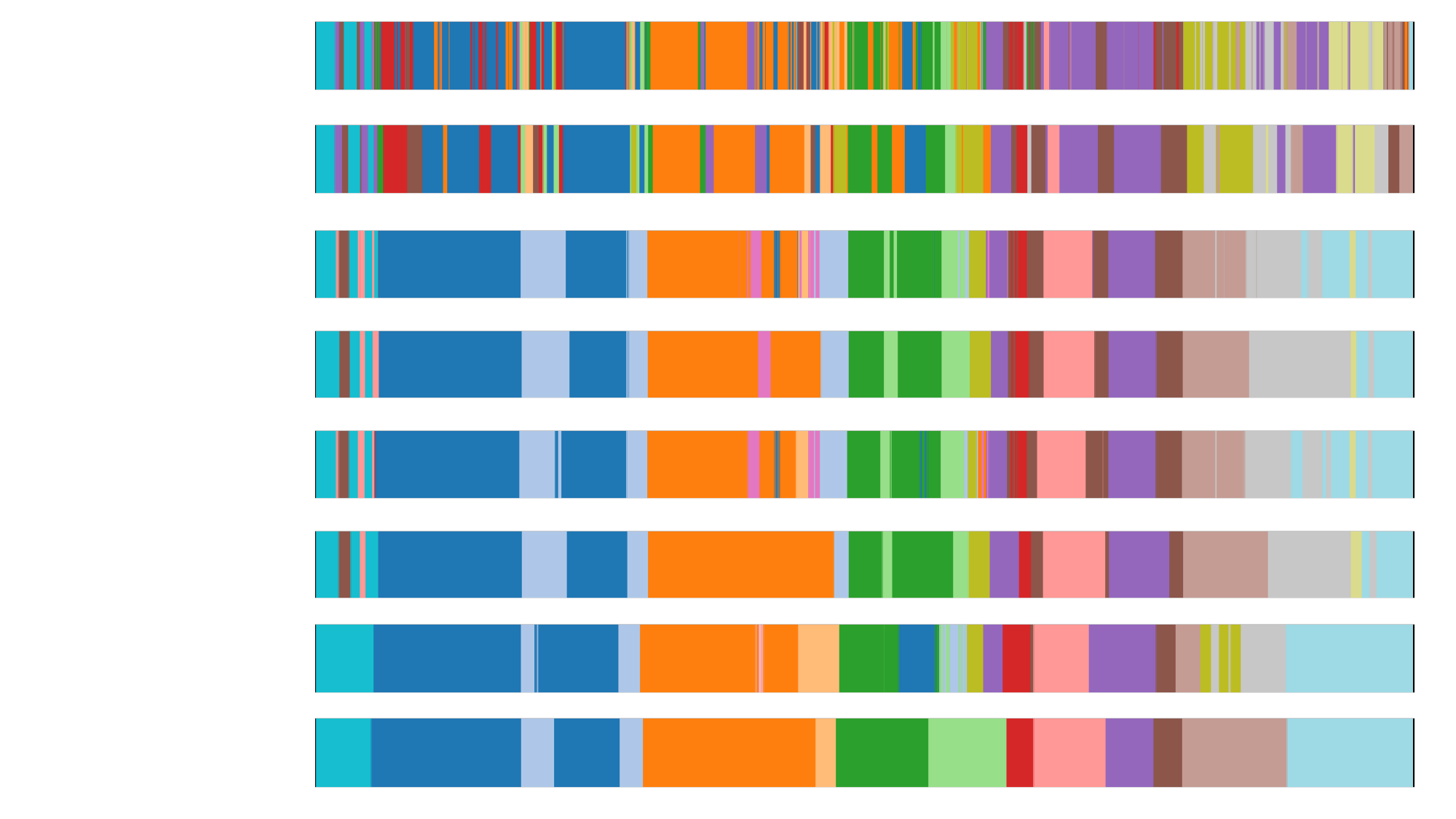}
    \put(7,51){\small LSTR~\cite{lstr}}
     \put(5.4,45.7){\small + post proc.}
    \put(8.4,36.8){\small Ours$^\text{online}$}
    \put(5.4,30.5){\small + post proc.}
    \put(9.2,23.3){\small Ours$^\text{semi}$}
    \put(5.4,17.1){\small + post proc.}
    \put(3,9.5){\small MS-TCN~\cite{ms-tcn}}
    \put(15.5,3.5){\small GT}
    \put(3,14){\tikz \draw[dashed,black, thick](10,10)--(21,10);}
    \put(3,27.8){\tikz \draw[dashed,black, thick](10,10)--(21,10);}
    \put(3,41.5){\tikz \draw[dashed,black, thick](10,10)--(21,10);}
    \put(36,2.0){\tikz \draw[draw=red,line width=0.5mm] (0,0) rectangle ++(0.4,5.9);}
    \put(87,2.0){\tikz \draw[draw=red,line width=0.5mm] (0,0) rectangle ++(0.75,5.9);}
    \put(78.5,2.0){\tikz \draw[draw=red,line width=0.5mm] (0,0) rectangle ++(0.35,5.9);}
    \put(23,2.0){\tikz \draw[draw=blue,line width=0.5mm] (0,0) rectangle ++(0.4,5.9);}
    \put(51,2.0){\tikz \draw[draw=blue,line width=0.5mm] (0,0) rectangle ++(0.6,5.9);}
    \put(60,2.0){\tikz \draw[draw=blue,line width=0.5mm] (0,0) rectangle ++(0.45,5.9);}
    \put(67,2.0){\tikz \draw[draw=blue,line width=0.5mm] (0,0) rectangle ++(0.4,5.9);}
\end{overpic}
\caption{Visualization of segmentation outputs for sequence ``rgb-01-1'' from 50Salads~\cite{50salads}. }
\label{fig:qual}
\end{figure}

\textbf{Qualitative Result.} 
\Cref{fig:qual} qualitatively compares the segmentation results from different approaches. It is clear to see that LSTR~\cite{lstr} suffers from the most prominent over-segmentation issue, which remains significant after the post-processing. Under the same configuration, our semi-online achieves slightly better results compared to the frame-by-frame online inference. Our post-processing, when applied, successfully removes the short fragments (blue boxes) in the raw prediction and refines the segmentation output.  However, it may reduce accuracy, particularly at action boundaries (red boxes).

For failure cases, we have the following two observations:
1) Detection of action start often delays due to the need for more frame information to predict new actions, especially when facing semantic ambiguities at action boundaries; 2) Persistent over-segmentation happens when the network makes incorrect but confident predictions, which could be improved with a stronger backbone or better temporal context modeling.

\begin{table}[t]
\centering
\begin{tabular}{llccccc}
\toprule
\multirow{3}{*}{{}}& \multirow{3}{*}{{Method}}& \multicolumn{5}{c}{{Breakfast~\cite{breakfast}}} \\ \cmidrule(lr){3-7}
 & & Acc & Edit & \multicolumn{3}{c}{F1 @ \{10, 25, 50\}} \\
\cmidrule(lr){1-2}\cmidrule(lr){3-7}
\multirow{3}{*}{\rotatebox[origin=c]{90}{offline}} & MS-TCN~\cite{ms-tcn} & 69.3 & 67.3 & 64.7 & 59.6 & 47.5\\
 & ASFormer~\cite{yi2021asformer} & 73.5 & 75.0 & 76.0 & 70.6 & 57.4 \\
 & DiffAct~\cite{liu2023diffusion} & 75.1 & 76.4 & 80.3 & 75.9 & 75.1 \\
\cmidrule(lr){1-2}\cmidrule(lr){3-7}
\multirow{7}{*}{\rotatebox[origin=c]{90}{online}} & MV-TAS~\cite{weakonline} & 41.6 & - & - & - & -  \\
 & LSTR~\cite{lstr} & 24.2 & 4.9 & 5.5 & 3.9 & 1.7  \\
 & Causal TCN & 55.3 & 18.7 & 15.1 & 11.7 & 8.3\\
 & Ours$^\text{online}$ & 56.7&19.3&16.8&13.9&9.3  \\
 & Ours$^\text{online}$ + p.p. & 52.9&55.7&54.8&45.8&30.5 \\
 & Ours$^\text{semi}$& \textbf{57.4} & 19.6 & 17.8 & 14.8 & 10.1  \\\rowcolor{gray!30}
 & Ours$^\text{semi}$ + p.p. & 53.8 & \textbf{57.5}& \textbf{56.4} & \textbf{47.3} & \textbf{31.4}  \\
\bottomrule
\end{tabular}
\caption{Comparison with the state-of-the-art methods on Breakfast.}
\label{tab:method_comparison_break}
\end{table}

\subsection{Runtime Analysis}

We evaluate the runtime performance of our approach using an Nvidia A40 GPU with both pre-computed I3D features and raw streaming RGB inputs, and present the inference times in~\Cref{tab:inf_time}. As shown, our approach can achieve up to 238.1 FPS when using pre-computed I3D features. To calculate the runtime for the entire segmentation pipeline, we take into consideration of the computational overheads of optical flow calculation and the I3D feature extraction. By leveraging a GPU backend for optical flow calculation, our full framework is able to achieve a runtime of 33.8 FPS. %

\textbf{Inference Latency.} The inference speed presented above is identical for both online and semi-online inference modes since their input sizes are the same. However, the latency can differ. In the online mode, inference is performed on a per-frame basis, meaning its latency is only dependent on the inference speed. In contrast, the semi-online mode incurs additional latency as it requires gathering frames up to the clip lengths before forming inputs.

Online inference offers better real-time responsiveness compared to semi-online inference, but the latter achieves superior performance as we discussed in~\Cref{subsec:abl}. The choice between these two modes depends on the application's priorities: if the real-time inference is critical, online inference is preferable; however, if accuracy is more important and the task is less time-sensitive,  semi-online inference is recommended.

\begin{table}[htb]
\centering
\begin{tabular}{lcccc}
\toprule
& {Ours  (I3D)}  &{OF  Comp.} & {I3D  Comp.}  &  {Ours (raw)}\\ \cmidrule(lr){1-1}\cmidrule(lr){2-2}\cmidrule(lr){3-5}
Time (ms) &\cellcolor{gray!30}{4.2}&4.8&20.5&\cellcolor{gray!30}{29.5}\\
FPS &\cellcolor{gray!30}{238.1}  &208.3 & 48.8 &\cellcolor{gray!30}{33.8} \\ 
\bottomrule
\end{tabular}
\caption{Runtime profile (in ms and FPS).}
\label{tab:inf_time}
\end{table}

\textbf{Limitation.} 
In this work, we only evaluate our approach on cooking videos, however, handling diverse and real-world videos may present several additional challenges. 
One common scenario involves interrupted actions, where a subject abruptly switches to a different action, leaving the ongoing action unfinished. These interruptions can be challenging for the model to handle effectively.
Additionally, the extended length of the video poses another challenge. Streaming videos can be infinitely long, so effectively managing and preserving long-form history within a fixed memory budget becomes a critical issue.

\section{Conclusion} \label{sec:limt}
This paper presents the first framework for the online segmentation of actions in procedural videos. Specifically, we propose an adaptive memory bank designed to accumulate and condense temporal context, alongside a feature augmentation module capable of injecting context information into inputs and producing enhanced representations. In addition, we propose a fast and effective post-processing technique aimed at mitigating the over-segmentation problem. Extensive experiments on common benchmarks have shown the effectiveness of our approach in addressing the online segmentation task.

\begin{ack}
This research is supported by the National Research Foundation, Singapore under its NRF Fellowship for AI (NRF-NRFFAI1-2019-0001). Any opinions, findings and conclusions or recommendations expressed in this material are those of the author(s) and do not reflect the views of National Research Foundation, Singapore.
\end{ack}
\bibliographystyle{abbrv}
\bibliography{main.bib}

\begin{thebibliography}{10}

\bibitem{ltcontext}
E.~Bahrami, G.~Francesca, and J.~Gall.
\newblock How much temporal long-term context is needed for action
  segmentation?
\newblock In {\em Proceedings of the IEEE/CVF International Conference on
  Computer Vision}, pages 10351--10361, 2023.

\bibitem{detec3}
S.~Cao, W.~Luo, B.~Wang, W.~Zhang, and L.~Ma.
\newblock E2e-load: end-to-end long-form online action detection.
\newblock In {\em Proceedings of the IEEE/CVF International Conference on
  Computer Vision}, pages 10422--10432, 2023.

\bibitem{i3d}
J.~Carreira and A.~Zisserman.
\newblock Quo vadis, action recognition? a new model and the kinetics dataset.
\newblock In {\em proceedings of the IEEE Conference on Computer Vision and
  Pattern Recognition}, pages 6299--6308, 2017.

\bibitem{gru}
J.~Chung, C.~Gulcehre, K.~Cho, and Y.~Bengio.
\newblock Empirical evaluation of gated recurrent neural networks on sequence
  modeling.
\newblock {\em arXiv preprint arXiv:1412.3555}, 2014.

\bibitem{detec2}
R.~De~Geest, E.~Gavves, A.~Ghodrati, Z.~Li, C.~Snoek, and T.~Tuytelaars.
\newblock Online action detection.
\newblock In {\em Computer Vision--ECCV 2016: 14th European Conference,
  Amsterdam, The Netherlands, October 11-14, 2016, Proceedings, Part V 14},
  pages 269--284. Springer, 2016.

\bibitem{survey}
G.~Ding, S.~Fadime, and A.~Yao.
\newblock Temporal action segmentation: An analysis of modern techniques.
\newblock {\em IEEE Transactions on Pattern Analysis and Machine Intelligence},
  2023.

\bibitem{ding2024coherent}
G.~Ding, H.~Golong, and A.~Yao.
\newblock Coherent temporal synthesis for incremental action segmentation.
\newblock In {\em Proceedings of the IEEE/CVF Conference on Computer Vision and
  Pattern Recognition}, pages 28485--28494, 2024.

\bibitem{ding2022leveraging}
G.~Ding and A.~Yao.
\newblock Leveraging action affinity and continuity for semi-supervised
  temporal action segmentation.
\newblock In {\em European Conference on Computer Vision}, pages 17--32.
  Springer, 2022.

\bibitem{ding2022temporal}
G.~Ding and A.~Yao.
\newblock Temporal action segmentation with high-level complex activity labels.
\newblock {\em IEEE Transactions on Multimedia}, 25:1928--1939, 2022.

\bibitem{unsup3}
Z.~Du, X.~Wang, G.~Zhou, and Q.~Wang.
\newblock Fast and unsupervised action boundary detection for action
  segmentation.
\newblock In {\em Proceedings of the IEEE/CVF Conference on Computer Vision and
  Pattern Recognition}, pages 3323--3332, 2022.

\bibitem{ms-tcn}
Y.~A. Farha and J.~Gall.
\newblock Ms-tcn: Multi-stage temporal convolutional network for action
  segmentation.
\newblock In {\em Proceedings of the IEEE/CVF conference on computer vision and
  pattern recognition}, pages 3575--3584, 2019.

\bibitem{gtea}
A.~Fathi, X.~Ren, and J.~M. Rehg.
\newblock Learning to recognize objects in egocentric activities.
\newblock In {\em CVPR 2011}, pages 3281--3288. IEEE, 2011.

\bibitem{weakonline}
R.~Ghoddoosian, I.~Dwivedi, N.~Agarwal, C.~Choi, and B.~Dariush.
\newblock Weakly-supervised online action segmentation in multi-view
  instructional videos.
\newblock In {\em Proceedings of the IEEE/CVF Conference on Computer Vision and
  Pattern Recognition}, pages 13780--13790, 2022.

\bibitem{guermal2024joadaa}
M.~Guermal, A.~Ali, R.~Dai, and F.~Br{\'e}mond.
\newblock Joadaa: joint online action detection and action anticipation.
\newblock In {\em Proceedings of the IEEE/CVF Winter Conference on Applications
  of Computer Vision}, pages 6889--6898, 2024.

\bibitem{minvis}
D.-A. Huang, Z.~Yu, and A.~Anandkumar.
\newblock Minvis: A minimal video instance segmentation framework without
  video-based training.
\newblock {\em Advances in Neural Information Processing Systems},
  35:31265--31277, 2022.

\bibitem{thumos}
H.~Idrees, A.~R. Zamir, Y.-G. Jiang, A.~Gorban, I.~Laptev, R.~Sukthankar, and
  M.~Shah.
\newblock The thumos challenge on action recognition for videos “in the
  wild”.
\newblock {\em Computer Vision and Image Understanding}, 155:1--23, 2017.

\bibitem{crf1}
Y.~Kong and Y.~Fu.
\newblock Human action recognition and prediction: A survey.
\newblock {\em International Journal of Computer Vision}, 130(5):1366--1401,
  2022.

\bibitem{breakfast}
H.~Kuehne, A.~Arslan, and T.~Serre.
\newblock {The language of actions: Recovering the syntax and semantics of
  goal-directed human activities}.
\newblock In {\em Proc. IEEE Conference on Computer Vision and Pattern
  Recognition (CVPR)}, pages 780--787, 2014.

\bibitem{unsuponline}
S.~Kumar, S.~Haresh, A.~Ahmed, A.~Konin, M.~Z. Zia, and Q.-H. Tran.
\newblock Unsupervised action segmentation by joint representation learning and
  online clustering.
\newblock In {\em Proceedings of the IEEE/CVF Conference on Computer Vision and
  Pattern Recognition}, pages 20174--20185, 2022.

\bibitem{fully1}
C.~Lea, M.~D. Flynn, R.~Vidal, A.~Reiter, and G.~D. Hager.
\newblock Temporal convolutional networks for action segmentation and
  detection.
\newblock In {\em proceedings of the IEEE Conference on Computer Vision and
  Pattern Recognition}, pages 156--165, 2017.

\bibitem{weak4}
J.~Li, P.~Lei, and S.~Todorovic.
\newblock Weakly supervised energy-based learning for action segmentation.
\newblock In {\em Proceedings of the IEEE/CVF international conference on
  computer vision}, pages 6243--6251, 2019.

\bibitem{weak1}
J.~Li and S.~Todorovic.
\newblock Action shuffle alternating learning for unsupervised action
  segmentation.
\newblock In {\em Proceedings of the IEEE/CVF Conference on Computer Vision and
  Pattern Recognition}, pages 12628--12636, 2021.

\bibitem{li2020ms}
S.~Li, Y.~A. Farha, Y.~Liu, M.-M. Cheng, and J.~Gall.
\newblock Ms-tcn++: Multi-stage temporal convolutional network for action
  segmentation.
\newblock {\em IEEE transactions on pattern analysis and machine intelligence},
  45(6):6647--6658, 2020.

\bibitem{robust}
S.~Lin, L.~Yang, I.~Saleemi, and S.~Sengupta.
\newblock Robust high-resolution video matting with temporal guidance.
\newblock In {\em Proceedings of the IEEE/CVF Winter Conference on Applications
  of Computer Vision}, pages 238--247, 2022.

\bibitem{liu2023diffusion}
D.~Liu, Q.~Li, A.-D. Dinh, T.~Jiang, M.~Shah, and C.~Xu.
\newblock Diffusion action segmentation.
\newblock In {\em International Conference on Computer Vision (ICCV)}, 2023.

\bibitem{swin}
Z.~Liu, Y.~Lin, Y.~Cao, H.~Hu, Y.~Wei, Z.~Zhang, S.~Lin, and B.~Guo.
\newblock Swin transformer: Hierarchical vision transformer using shifted
  windows.
\newblock In {\em Proceedings of the IEEE/CVF international conference on
  computer vision}, pages 10012--10022, 2021.

\bibitem{fully2}
D.~Moltisanti, S.~Fidler, and D.~Damen.
\newblock Action recognition from single timestamp supervision in untrimmed
  videos.
\newblock In {\em Proceedings of the IEEE/CVF Conference on Computer Vision and
  Pattern Recognition}, pages 9915--9924, 2019.

\bibitem{otalc}
M.~K. Myers, N.~Wright, A.~S. McGough, and N.~Martin.
\newblock O-talc: Steps towards combating oversegmentation within online action
  segmentation.
\newblock {\em arXiv preprint arXiv:2404.06894}, 2024.

\bibitem{weak2}
R.~Rahaman, D.~Singhania, A.~Thiery, and A.~Yao.
\newblock A generalized and robust framework for timestamp supervision in
  temporal action segmentation.
\newblock In {\em European Conference on Computer Vision}, pages 279--296.
  Springer, 2022.

\bibitem{weak5}
A.~Richard, H.~Kuehne, and J.~Gall.
\newblock Action sets: Weakly supervised action segmentation without ordering
  constraints.
\newblock In {\em Proceedings of the IEEE conference on Computer Vision and
  Pattern Recognition}, pages 5987--5996, 2018.

\bibitem{weak3}
A.~Richard, H.~Kuehne, A.~Iqbal, and J.~Gall.
\newblock Neuralnetwork-viterbi: A framework for weakly supervised video
  learning.
\newblock In {\em Proceedings of the IEEE conference on Computer Vision and
  Pattern Recognition}, pages 7386--7395, 2018.

\bibitem{unsup2}
S.~Sarfraz, N.~Murray, V.~Sharma, A.~Diba, L.~Van~Gool, and R.~Stiefelhagen.
\newblock Temporally-weighted hierarchical clustering for unsupervised action
  segmentation.
\newblock In {\em Proceedings of the IEEE/CVF Conference on Computer Vision and
  Pattern Recognition}, pages 11225--11234, 2021.

\bibitem{sener2020temporal}
F.~Sener, D.~Singhania, and A.~Yao.
\newblock Temporal aggregate representations for long-range video
  understanding.
\newblock In {\em Computer Vision--ECCV 2020: 16th European Conference,
  Glasgow, UK, August 23--28, 2020, Proceedings, Part XVI 16}, pages 154--171.
  Springer, 2020.

\bibitem{unsup1}
F.~Sener and A.~Yao.
\newblock Unsupervised learning and segmentation of complex activities from
  video.
\newblock In {\em Proceedings of the IEEE Conference on Computer Vision and
  Pattern Recognition}, pages 8368--8376, 2018.

\bibitem{rule1}
Z.~Shou, D.~Wang, and S.-F. Chang.
\newblock Temporal action localization in untrimmed videos via multi-stage
  cnns.
\newblock In {\em Proceedings of the IEEE conference on computer vision and
  pattern recognition}, pages 1049--1058, 2016.

\bibitem{semi1}
D.~Singhania, R.~Rahaman, and A.~Yao.
\newblock Iterative contrast-classify for semi-supervised temporal action
  segmentation.
\newblock {\em Proceedings of the AAAI Conference on Artificial Intelligence},
  36(2):2262–2270, Jul 2022.

\bibitem{singhania2023c2f}
D.~Singhania, R.~Rahaman, and A.~Yao.
\newblock C2f-tcn: A framework for semi-and fully-supervised temporal action
  segmentation.
\newblock {\em IEEE Transactions on Pattern Analysis and Machine Intelligence},
  2023.

\bibitem{50salads}
S.~Stein and S.~J. McKenna.
\newblock {Combining embedded accelerometers with computer vision for
  recognizing food preparation activities}.
\newblock In {\em Proceedings of the 2013 ACM international joint conference on
  Pervasive and ubiquitous computing}, pages 729--738, 2013.

\bibitem{vaswani2017attention}
A.~Vaswani, N.~Shazeer, N.~Parmar, J.~Uszkoreit, L.~Jones, A.~N. Gomez,
  {\L}.~Kaiser, and I.~Polosukhin.
\newblock Attention is all you need.
\newblock {\em Advances in neural information processing systems}, 30, 2017.

\bibitem{rule2}
S.~Venkatesh, D.~Moffat, and E.~R. Miranda.
\newblock Investigating the effects of training set synthesis for audio
  segmentation of radio broadcast.
\newblock {\em Electronics}, 10(7):827, 2021.

\bibitem{unders1}
J.~Wang, G.~Chen, Y.~Huang, L.~Wang, and T.~Lu.
\newblock Memory-and-anticipation transformer for online action understanding.
\newblock In {\em Proceedings of the IEEE/CVF International Conference on
  Computer Vision}, pages 13824--13835, 2023.

\bibitem{detec1}
X.~Wang, S.~Zhang, Z.~Qing, Y.~Shao, Z.~Zuo, C.~Gao, and N.~Sang.
\newblock Oadtr: Online action detection with transformers.
\newblock In {\em Proceedings of the IEEE/CVF International Conference on
  Computer Vision}, pages 7565--7575, 2021.

\bibitem{xu2019temporal}
M.~Xu, M.~Gao, Y.-T. Chen, L.~S. Davis, and D.~J. Crandall.
\newblock Temporal recurrent networks for online action detection.
\newblock In {\em Proceedings of the IEEE/CVF international conference on
  computer vision}, pages 5532--5541, 2019.

\bibitem{lstr}
M.~Xu, Y.~Xiong, H.~Chen, X.~Li, W.~Xia, Z.~Tu, and S.~Soatto.
\newblock Long short-term transformer for online action detection.
\newblock {\em Advances in Neural Information Processing Systems},
  34:1086--1099, 2021.

\bibitem{yi2021asformer}
F.~Yi, H.~Wen, and T.~Jiang.
\newblock Asformer: Transformer for action segmentation.
\newblock In {\em BMVC}, 2021.

\bibitem{ctvis}
K.~Ying, Q.~Zhong, W.~Mao, Z.~Wang, H.~Chen, L.~Y. Wu, Y.~Liu, C.~Fan,
  Y.~Zhuge, and C.~Shen.
\newblock Ctvis: Consistent training for online video instance segmentation.
\newblock In {\em Proceedings of the IEEE/CVF International Conference on
  Computer Vision}, pages 899--908, 2023.

\bibitem{dvis}
T.~Zhang, X.~Tian, Y.~Wu, S.~Ji, X.~Wang, Y.~Zhang, and P.~Wan.
\newblock Dvis: Decoupled video instance segmentation framework.
\newblock In {\em Proceedings of the IEEE/CVF International Conference on
  Computer Vision}, pages 1282--1291, 2023.

\bibitem{hacs}
H.~Zhao, A.~Torralba, L.~Torresani, and Z.~Yan.
\newblock Hacs: Human action clips and segments dataset for recognition and
  temporal localization.
\newblock In {\em Proceedings of the IEEE/CVF International Conference on
  Computer Vision}, pages 8668--8678, 2019.

\end{thebibliography}

\clearpage
\appendix

\section{Appendix / supplemental material}
\subsection{Standard vs. Causal Convolution} 

\begin{figure*}[t]
    \centering
    \includegraphics[width=0.8\linewidth]{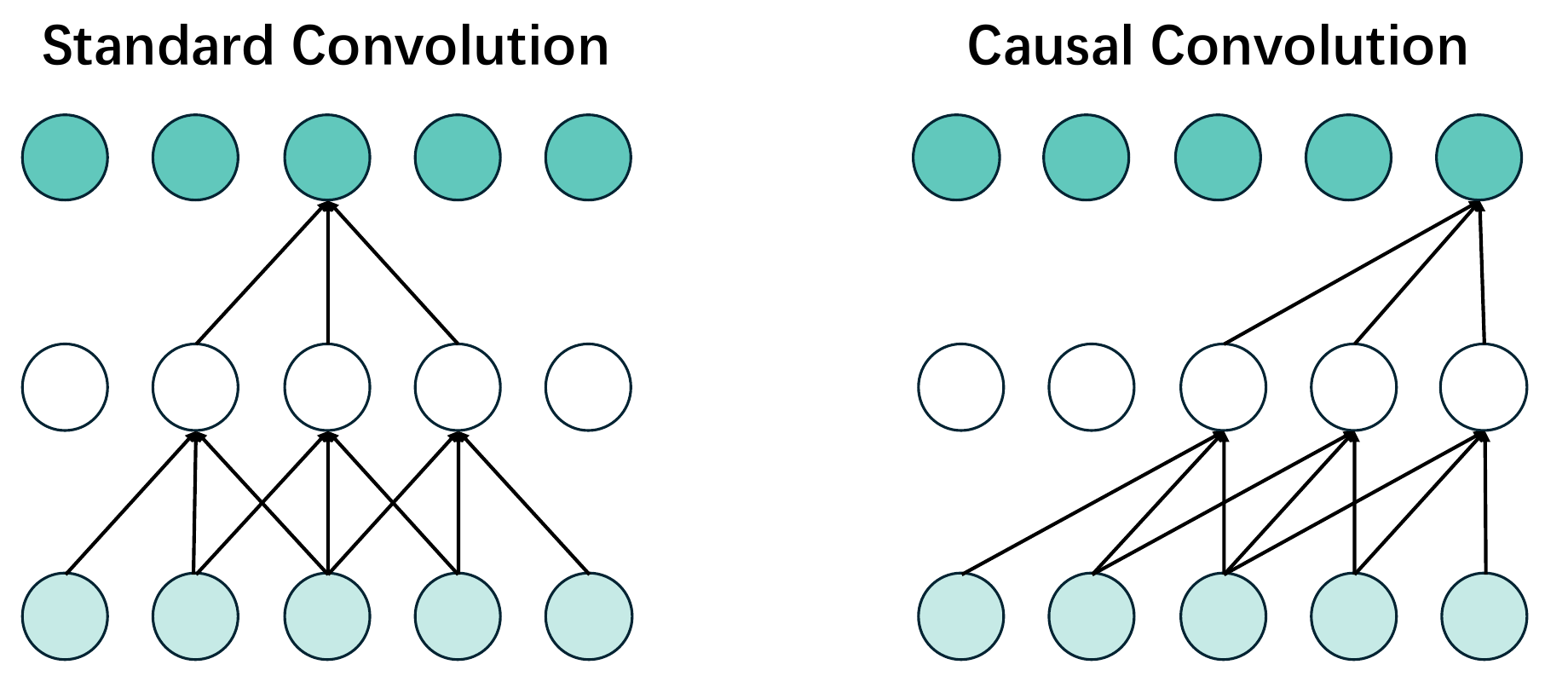}
\caption{Standard vs. Causal Convolution}
\label{fig:conv}
\end{figure*}

\textbf{Standard convolution.} As shown in \Cref{fig:conv} (left), the receptive field of a standard convolution includes both past and future inputs. Mathematically, for an input sequence $x(t)$ and a filter $h(k)$, the output $y(t)$ at time $t$ is given by:
\begin{equation}
y(t) = \sum_{k=-K}^{K} h(k) \cdot x(t-k)
\end{equation}
where $K$ is the size of the filter. This means the output at time $t$ depends on inputs from $t-K$ to $t+K$.

\textbf{Causal convolution.} As shown in \Cref{fig:conv} (right), the receptive field of a causal convolution includes only the past and current inputs, ensuring that the output at time $t$ does not depend on future inputs. Mathematically, the output $y(t)$ is given by:
\begin{equation}
y(t) = \sum_{k=0}^{K} h(k) \cdot x(t-k) 
\end{equation}
where $K$ is the size of the filter. This means the output at time $t$ depends only on inputs from $t$ to $t-K$.

\subsection{CFA Formula}
The attention mechanism~\cite{vaswani2017attention} is written as:
\begin{equation}
\text{Attention}(Q, K, V) = \text{SoftMax}\left( \frac{Q \times K^T}{\sqrt{d}} \right) \times V
\end{equation}
where $Q,K,V$ represents  query,  key and  value, respectively,  and  \(d\) is the hidden dimension.

We use a Transformer decoder \cite{vaswani2017attention} to obtain the memory encoding, and TransDecoder() is formulated as follows:
\begin{equation}
\tilde{M}_{k-1}^{\text{TD}}\ = \text{SelfAttn}(M_{k-1}, M_{k-1}, M_{k-1}) + \text{CrossAttn}(M_{k-1}, c^{\text{GRU}}_k, c^{\text{GRU}}_k) + \text{FFN}
\end{equation}
Here, the FFN (Feed-Forward Network) is a two-layer fully connected network, \(M_{k-1}\) is the memory bank, which first undergoes self-attention. The output of the self-attention mechanism is used as the query for cross-attention, where \(c^{\text{GRU}}_k\) serves as the key and value. This interaction results in a more effective memory encoding \(\tilde{M}_{k-1}^{\text{TD}}\).

We split the input feature of size $C \times T$ to 2 windows with size $C \times \frac{T}{2}$, and perform Swin \cite{swin} self attention within each local window independently, and Cross attention for every two consecutive local windows. The Swin attention mechanism can be formulated as:
\begin{equation} \label{eql:swinattn}
\text{Swin Attention}(Q, K, V) = \text{SoftMax}\left( \frac{Q \times K^T}{\sqrt{d}} + B \right) \times V
\end{equation}
Where the $B$ is the relative position of the window. Then, our method produces context-augmented features $\tilde{c}_k$ using \cref{eql:self} and \cref{eql:cross}

Our self-attention based on \cref{eql:swinattn}:
\begin{equation}
\text{SelfAttn}(c_k^{GRU}, c_k^{GRU}, c_k^{GRU}) = \text{SoftMax}\left( \frac{c_k^{GRU} \times (c_k^{GRU})^T}{\sqrt{d}} + B \right) \times c_k^{GRU}
\end{equation}
where $c$ as the clip features and $k$ as the current step, $c_k$ passes through a GRU to obtain $c_k^{\text{GRU}}$.
Next, we use cross-attention to interact with the output of the self-attention 
$c_k^{\text{SA}}$ with the output memory bank of the Transformer decoder, \(\tilde{M}_{k-1}^{\text{TD}}\):
\begin{equation}
\text{CrossAttn}(c_k^{\text{SA}}, \Tilde{M}_{k-1}^{\text{TD}},\tilde{M}_{k-1}^{\text{TD}}) =  \text{SoftMax}\left( \frac{c_k^{\text{SA}} \times (\tilde{M}_{k-1}^{\text{TD}})^T}{\sqrt{d}} + B \right) \times \tilde{M}_{k-1}^{\text{TD}}
\end{equation}

\subsection{Implementation}
\textbf{Hyper-parameters.} As shown in \cref{tab:hyperparameters}, the hyper-parameter settings are generally the same for each dataset. GTEA uses a shorter window size because the longest video is only about 2000 frames, whereas the other datasets all use a window size of 128. The optimal confidence threshold for segmental metrics varies for each dataset: for GTEA is 0.6; for 50Salads is 0.9; and for Breakfast is 0.8.

\begin{table}[h]
\centering
\begin{tabular}{lccc}
\toprule
\textbf{Hyper-parameters} & \textbf{GTEA} & \textbf{50Salads} & \textbf{Breakfast} \\
\midrule
Learning rate & $5e^{4}$ & $5e^{4}$ & $5e^{4}$ \\
Epochs & 50 & 50 & 50 \\
No. GRU layers & 1 & 1 & 1 \\\rowcolor{gray!30}
$w$ & 64 & 128 & 128 \\
$\sigma$ & 1/16 & 1/16 & 1/16 \\\rowcolor{gray!30}
$\theta$ & 0.6 & 0.9 & 0.8 \\
iteration $I$ & 2 & 2 & 2 \\
TD heads & 8 & 8 & 8 \\
SwinAttn heads & 4 & 4 & 4 \\
No. Causal TCN stages & 1 & 1 & 1 \\
No. Causal dilated Conv layers & 10 & 10 & 10 \\
\bottomrule
\end{tabular}
\caption{Hyper-parameter settings for GTEA, 50Salads, and Breakfast datasets.}
\label{tab:hyperparameters}
\end{table}

\subsection{AsFormer Performance}

\begin{table}[t]
\centering
\scalebox{0.8}{
\begin{tabular}{lccccccccccccccccc}
\toprule
\multirow{4}{*}{} & \multirow{4}{*}{{p.p.}} & \multicolumn{5}{c}{{GTEA}} & \multicolumn{5}{c}{{50Salads}}&\multicolumn{5}{c}{{Breakfast}} \\ 
\cmidrule(lr){3-7} \cmidrule(lr){8-12}\cmidrule(lr){13-17}
& & Acc & Edit & \multicolumn{3}{c}{F1 @ \{10, 25, 50\}} & Acc & Edit & \multicolumn{3}{c}{F1 @ \{10, 25, 50\}} & Acc & Edit & \multicolumn{3}{c}{F1 @ \{10, 25, 50\}} \\ 
\cmidrule(lr){1-2}\cmidrule(lr){3-7} \cmidrule(lr){8-12}\cmidrule(lr){13-17}

Offline& - &79.7 & 84.6 & 90.1 & 88.8 & 79.2 & 85.6 & 79.6 & 85.1 & 83.4 & 76.0 & 73.5 & 75.0 & 76.0 & 70.6 & 57.4\\
\cmidrule(lr){1-2}\cmidrule(lr){3-7} \cmidrule(lr){8-12}\cmidrule(lr){13-17}
& - &75.0 & 69.7 & 77.7 & 74.0 & 62.0 & 77.5 & 29.1 & 37.9 & 35.3 & 28.6 &64.9 &32.1&31.2&27.4&20.2\\ %
\multirow{-2}{*}{{Online}}& \checkmark &72.8 & 77.8 & 84.5 & 80.8 & 64.2 & 69.4 & 36.4 & 70.6 & 65.7 & 52.3 &63.1 &60.1 &61.6 &54.3 &39.2\\
\cmidrule(lr){1-2}\cmidrule(lr){3-7} \cmidrule(lr){8-12}\cmidrule(lr){13-17}
& - & 76.5 & 71.3 & 79.0 & 76.7 & 63.1 & 78.5 & 29.7 & 38.5 & 36.2 & 30.4 &64.8&37.0&33.9&30.0&22.6\\ %
 \multirow{-2}{*}{Semi}& \checkmark & 74.7 & 79.6 & 86.3 & 82.8 & 67.0 & 71.0 & 64.9 & 72.2 & 67.2 & 53.8 &64.0&63.2&64.9&57.5&43.0\\ %

\bottomrule
\end{tabular}}
\caption{
Performance of our approach when using ASFormer as the segmentation backbone. 
}
\label{tab:asformer}
\end{table}

We conduct experiments on three common TAS datasets, where we replace the MS-TCN backbone with AsFormer. In MS-TCN, the transition to an online method is relatively straightforward, as it only requires replacing all the standard convolution layers with causal convolution layers. However, in AsFormer, the transformation involves more extensive modifications. In addition to replacing the convolution layers with causal convolutions, we also modify the standard attention layers into causal attention layers. Furthermore, we incorporate our proposed GRU, CFA, Memory Bank, and a Post-processing module to ensure that AsFormer transitions from an offline method to an online method. Our approach remains highly effective in boosting online segmentation performance while maintaining the strengths of the AsFormer architecture.

\end{document}